\documentclass[english]{article}
\usepackage[T1]{fontenc}
\usepackage[latin9]{inputenc}
\usepackage{babel}
\usepackage{geometry}
\usepackage{graphicx}
\usepackage{biblatex}
\usepackage{algorithm}
\usepackage{algpseudocode}
\usepackage{amsmath}
\usepackage{enumitem}
\usepackage{csquotes}
\usepackage{verbatim}
\usepackage{subcaption}
\usepackage{hyperref}
\usepackage{tikz}
\usetikzlibrary{decorations.pathreplacing,calc}
\newcommand{\tikzmark}[1]{\tikz[overlay,remember picture] \node (#1) {};}

\newcommand*{\AddNote}[4]{%
    \begin{tikzpicture}[overlay, remember picture]
        \draw [decoration={brace,amplitude=0.4em},decorate, thick,black]
            ($(#3)!(#1.south)!($(#3)-(0,1)$)$) --  
            ($(#3)!(#2.south)!($(#3)-(0,1)$)$)
                node [align=center, text width=2.5cm, pos=0.5, anchor=west] {#4};
    \end{tikzpicture}
}%

\begin{document}

\title{LATE Ain'T Earley: A Faster Parallel Earley Parser}
\author{Peter Ahrens\\\texttt{pahrens@mit.edu} \and John Feser\\\texttt{feser@mit.edu} \and Joseph Hui\\\texttt{ctunoku@mit.edu}}

\maketitle

\begin{abstract}
  We present the LATE algorithm, an asynchronous variant of the Earley algorithm for parsing context-free grammars. The Earley algorithm is naturally task-based, but is difficult to parallelize because of dependencies between the tasks. We present the LATE algorithm, which uses additional data structures to maintain information about the state of the parse so that work items may be processed in any order. This property allows the LATE algorithm to be sped up using task parallelism. We show that the LATE algorithm can achieve a 120x speedup over the Earley algorithm on a natural language task.
\end{abstract}

\section{Introduction}

    
    
    Improvements in the efficiency of parsers for context-free grammars (CFGs) have the potential to speed up applications in software development, computational
linguistics, and human-computer interaction. The Earley parser has an asymptotic complexity that scales with the complexity of the CFG, a unique, desirable trait among parsers for arbitrary CFGs. However, while the more commonly used Cocke-Younger-Kasami (CYK)~\cite{cocke_programming_1970,kasami_efficient_1966,younger_recognition_1967} parser has been successfully parallelized~\cite{canny_multi-teraflop_2013,mark_johnson_parsing_2011}, the Earley algorithm has seen relatively few attempts at parallelization. Our research objectives were to understand when there exists parallelism in the Earley algorithm, and to explore methods for exploiting this parallelism.

We first tried to naively parallelize the Earley algorithm by processing the
Earley items in each Earley set in parallel. We found that this approach does
not produce any speedup, because the dependencies between Earley items force
much of the work to be performed sequentially. In this paper, we discuss the
modifications that we made to the Earley algorithm to address this problem. We
add additional data structures that allow us to remove the sequential
dependencies between the Earley sets. Our new algorithm (Sec.~\ref{sec:late}),
LATE (LATE Ain'T Earley), can process Earley items added to its global work queue
in parallel in any order. This property allows us to explore and exploit the
opportunities for parallelism in the Earley algorithm. We implemented the LATE
parser using shared-memory parallelism (Sec.~\ref{sec:impl}) and tested its
performance on three datasets (Sec.~\ref{sec:exp}). LATE achieved significant
speedups on certain datasets, but not others; in particular, it is much more
efficient than the Earley algorithm for grammars that are very ambiguous, but
less efficient for grammars that are unambiguous (Sec.~\ref{sec:res}).


\section{Motivation}



Context-free grammars (CFGs) have a wide range of uses, from parsing computer
programs to analyzing natural language. Although efficient algorithms exist
for many specialized classes of grammars, efficient algorithms for parsing
arbitrary CFGs, such as the Earley algorithm or the Cocke-Younger-Kasami (CYK)
algorithm, are highly desirable for two reasons. First, applications in
linguistics require parsers to handle ambiguous grammars, because natural
language is inherently ambiguous. Second, parsers for restricted grammar classes
can be difficult to use, because they require the user to rewrite their grammar
so that it fits in the parser's accepted class.


The Earley algorithm is a good candidate for an efficient general-purpose CFG parser. It has good computational complexity, running in $O(n^3)$ time in general, $O(n^2)$ time for unambiguous grammars, and in $O(n)$ time for a class which includes most LR($k$) grammars~\cite{earley_efficient_1970}. In particular, the asymptotic behavior of the Earley algorithm compares favorably to that of the CYK algorithm, which takes $O(n^3)$ time in all cases~\cite{cocke_programming_1970,kasami_efficient_1966,younger_recognition_1967}.


Exploiting parallelism is an important part of making an algorithm perform well
on modern hardware. Parsers are challenging to parallelize because they contain
complex serial dependencies. However, the Earley and CYK algorithms compute not
just a single parse, but the set of all parses of the input. The set of all
parses can be large if the grammar is ambiguous, so there is a clear opportunity
for parallelism. Previous work shows that the CYK algorithm can be parallelized
successfully~\cite{mark_johnson_parsing_2011}. The CYK algorithm is a dynamic programming algorithm which
computes a matrix that represents the set of all parses. This matrix can be
computed in parallel using a wavefront method. Parallelization attempts have
yielded up to a 60x speedup over the sequential
algorithm~\cite{mark_johnson_parsing_2011}. The Earley algorithm has a more
complex structure, with several branches in the main loop, and it relies on a
complex set-based data structure. Attempts to parallelize the Earley algorithm
have been less successful, and rely on introducing speculative work items,
yielding a maximum speedup of 5.44x~\cite{fowler_parallel_2009}. Although there are
many performance engineering techniques one might apply to the Earley algorithm, we limit our focus to parallel execution since it has proven to be the most challenging technique to apply in this case.


Despite the difficulty involved in parallelizing parsers, some theoretical work gives us faith that there exists usable parallelism in the Earley parser. Context-free grammar parsing can be reduced to binary matrix
multiplication~\cite{valiant_general_1975}, which proves the existence of an $O(n^\omega)$ algorithm for context-free parsing, where $\omega = 2.37$ is the
best known exponent in a matrix multiplication algorithm (that is, the best known matrix multiplication algorithm also runs in $O(n^\omega)$ time). Boolean matrix
multiplication can also be reduced to context-free grammar
parsing~\cite{lee_fast_2002}. Boolean matrix multiplication is highly
parallelizable, so it seems likely that a straightforward, highly parallel
CFG parsing algorithm exists as well.





\section{Previous Work}

There is significant existing work on speeding up the CYK algorithm using multiple CPUs or GPUs. Less attention has been paid to the Earley algorithm, because it has a more complex implementation.

\Citeauthor{mark_johnson_parsing_2011} parallelizes the CYK algorithm for probabilistic context-free grammars (PCFGs) using Symmetric Multi-Processing (SMP) parallelism on CPUs and Single-Instruction Multiple-Thread (SIMT) parallelism on GPUs, achieving a 60x speedup in some cases~\cite{mark_johnson_parsing_2011}. The CYK algorithm for PCFGs consists of a number of nested loops with a basic arithmetic operation (similar to matrix multiplication), and the main work of the algorithm can be split into many chunks that can be processed independently, allowing a large speedup with a relatively straightforward implementation.

\Citeauthor{canny_multi-teraflop_2013} discuss several techniques that improve
the performance of PCFG parsing on the GPU using the CYK
algorithm~\cite{canny_multi-teraflop_2013}. They note that the basic arithmetic
step of the CYK algorithm is bottlenecked by the speed of the various kinds of
memory (e.g. shared memory, constant memory, and L2 cache). By slightly
adjusting the structure of the algorithm to take advantage of compiler
optimizations, they allow almost all of the computations to take place in a
small number of register variables (the fastest type of memory). This allows CYK
parsing to proceed at close to the maximum speed of the GPU. That is, the GPU is almost always performing useful computation instead of waiting for memory reads/writes.

Although there are several theoretical parallel Earley parsers, there are few
practical attempts. In order to allow work to be performed in parallel,
\Citeauthor{fowler_parallel_2009} divide the work items into independent blocks
by how far back their rule began~\cite{fowler_parallel_2009}. However, the
blocks are not truly independent, because the parser state at the beginning of a
block depends on the state at the end of the previous block. A speculation step
is introduced to break this dependency and allow blocks to be processed in
parallel, which increases the amount of work that the parser performs. The cost
of speculation increases as the number of blocks increases, which means that the
returns of parallelism diminish as the degree of parallelism increases. This
method achieves a maximum speedup of 5.44x over their single-threaded
implementation using 8 threads when parsing Java source code.

\section{Methodology and Implementation}

Our goal was to understand the available parallelism in the Earley algorithm, and to characterize the grammars which can be effectively parsed in parallel. Since the Earley algorithm uses several global data structures, we focused on a multicore model so that we could take advantage of shared memory. We started by implementing our algorithms in Python, a high-level dynamic programming language, which allowed us to test new features quickly and gave us a reference implementation to compare against. We then created a C++ implementation in order to conduct a performance analysis.

We started with a naive parallelization of the Earley algorithm by processing each Earley set in parallel. However, we found that this approach did not provide any speedup since the communication overhead was too large relative to the amount of parallel work (Sec.~\ref{sec:parallel}).

After analyzing the Earley algorithm more closely, we found a way to modify the algorithm by adding additional data structures, making it more amenable to parallelization without sacrificing correctness. This new algorithm does the same work as the Earley algorithm, but uses three new data structures to keep track of which steps have already been executed, allowing work items to be processed out of order.

We named our modified Earley algorithm the LATE (LATE Ain'T Earley) algorithm. We validated LATE theoretically by showing that it performs the same actions as the Earley algorithm (Sec.~\ref{sec:late}). LATE freed us to process work items in parallel as soon as they were discovered, which allowed us to better explore the available parallelism in the Earley algorithm (Sec.~\ref{sec:late-parallel}). We validated LATE experimentally by comparing the output of our LATE implementations to that of a reference Earley implementation. We then created a C++ implementation of the LATE algorithm (both serial and parallel) in order to measure its performance against the Earley algorithm (Sec.~\ref{sec:impl}). We considered linguistic principles such as the ambiguity of the grammar in our analysis and selection of the different inputs to our algorithms. 

In this section, we discuss the Earley algorithm, starting with the standard serial implementation and a naive parallelization. We then present LATE, a novel parsing algorithm created by augmenting the Earley algorithm with additional data structures to allow for asynchronous execution of Earley tasks.


\begin{algorithm}
  \begin{algorithmic}[1]
  \Function{Scanner}{$(S \rightarrow \alpha \bullet T \beta, i), k$}\label{ln:scan-start}
  \If{$W[k] = T$}
  \State $C[k+1] \gets C[k+1] \cup (S \rightarrow \alpha T \bullet \beta, i)$
  \EndIf
  \EndFunction\label{ln:scan-end}
  \State
  \Function{Predictor}{$(S \rightarrow \alpha \bullet N \beta, i), k$}\label{ln:pred-start}
  \For{$N \rightarrow \gamma \in G$}
  \State $C[k] \gets C[k] \cup (N \rightarrow \bullet \gamma, k)$
  \EndFor
  \EndFunction\label{ln:pred-end}
  \State
  \Function{Completer}{$(S \rightarrow \alpha \bullet, i), k$}\label{ln:cmp-start}
  \For{$(T \rightarrow \beta \bullet S \gamma, j) \in C[i]$}
  \State $C[k] \gets C[k] \cup (T \rightarrow \beta S \bullet \gamma, j)$
  \EndFor
  \EndFunction\label{ln:cmp-end}
  \State
  \Function{Earley}{$G, W$}
  \State Let $C$ be an array of $|W|$ sets. \label{ln:chart}
  \For{$\text{START} \rightarrow \beta \in G$}\label{ln:init-start}
  \State $C[0] \gets C[0] \cup (\text{START} \rightarrow \bullet \beta, 0)$
  \EndFor\label{ln:init-end}
  \For{$k$ from $1$ to $|W|$}
  \ForAll{States $s \in C[k]$}\label{ln:inner-start}
  \If{$s = (S \rightarrow \alpha \bullet, i)$} \Call{Completer}{$s, k$}
  \ElsIf{$s = (S \rightarrow \alpha \bullet T \beta, i)$ and $T$ is a
    non-terminal} \Call{Predictor}{$s, k$}
  \Else{\ \Call{Scanner}{$s, k$}}
  \EndIf
  \EndFor\label{ln:inner-end}
  \EndFor
  \State \Return $(\text{START} \rightarrow \alpha \bullet, i) \in C[|W|]$
  \EndFunction
\end{algorithmic}


  \label{alg:earley}
  \caption{The serial Earley algorithm for parsing CFGs~\cite{earley_efficient_1970}.}
\end{algorithm}

\subsection{Serial Earley Parsing}\label{sec:serial}
The Earley algorithm~\cite{earley_efficient_1970} is a type of chart parser. It constructs a data structure which contains all of the parses of the input sentence in a compressed form. This data structure is an array containing \emph{Earley sets}, with one set for each symbol in the input string. Each set contains \emph{Earley items}. An Earley item represents a partial parse of some rule. Earley items consist of:
\begin{itemize}
\item \emph{A grammar rule:} Grammar rules are of the form $S \to \alpha \beta$. We use $\alpha, \beta, \gamma$ to represent strings of symbols and $S, T, N$ to represent individual symbols.
\item \emph{A position in the rule:} The position describes where in the rule the parser currently is.
\item \emph{An origin position:} The origin position is the position in the input string where the rule started parsing.
\item \emph{A current position:} Each Earley item is associated with a position in the input string, representing the amount of the string that has been seen by the parser.
\end{itemize}
We write Earley items as 3-tuples $(S \to \alpha \bullet \beta, i, k)$, where $i$ is the origin of the rule and $k$ is the current position. We use a $\bullet$ to represent the current parse position within the rule itself.  Thus, this Earley item represents a partial parse of the nonterminal $S$ using input symbols from positions $i$ to $k$, where $\alpha$ has already been parsed and we have yet to parse $\beta$. When $k$ is clear from the context we may abbreviate this item as $(S \to \alpha \bullet \beta, i)$.

The Earley chart is a data structure which stores several sets of Earley items (one set per input position). The Earley set $k$ consists of all Earley items with current position $k$. The Earley algorithm uses the START rule to initialize the chart (Lines~\ref{ln:init-start}-\ref{ln:init-end}). The main loop iterates through the chart and applies one of three subroutines for each new item added to a set:
\begin{itemize}
  \item \textsc{Scanner}: The scanner runs on unfinished rules of the form $(S \to \alpha \bullet T \beta, i)$ where $T$ is a terminal. If the string contains $T$ in position $k$, the scanner advances the rule and puts it in the next Earley set (Lines~\ref{ln:scan-start}-\ref{ln:scan-end}). 
  \item \textsc{Predictor}: The predictor runs on unfinished rules of the form $(S \to \alpha \bullet N \beta, i)$ where $N$ is a nonterminal. For these rules to advance, $N$ needs to be parseable. The predictor inserts all of the grammar rules which can parse $N$ into the current Earley set (Lines~\ref{ln:pred-start}-\ref{ln:pred-end}). The advancing of rules in the scanner and predictor is analogous to the shift action in an LR parser.
  \item \textsc{Completer}: The completer runs on finished Earley items, where the entire right hand side of the rule has been parsed. If the rule is finished, then we have seen a successful parse of the nonterminal on the left hand side, and we need to advance rules that were waiting on these parses.  When it finds a finished rule $S \rightarrow \alpha \bullet$, the completer returns to the Earley set where the rule originated. Any rule that has $S$ as its next symbol to parse can be advanced, so the completer puts them in the current item set (Lines~\ref{ln:cmp-start}-\ref{ln:cmp-end}). The operation of the completer is analogous to the reduce action in an LR parser.
\end{itemize}

\subsection{Naive Parallel Earley Parsing}\label{sec:parallel}
A simple approach to parallelizing the Earley algorithm runs the inner loop (Lines~\ref{ln:inner-start}-\ref{ln:inner-end}) using several worker threads, treating the set of Earley items like a work queue. Unfortunately, it is difficult to make this parallel processing efficient, because we need to ensure that each thread does enough work to balance the overhead of distributing the work between threads. The Earley sets are often small, meaning that there is little parallel work. Also, the threads must be synchronized after completing the processing of each Earley set. When we investigated this technique, we found that the parallel overhead is greater than the speedup from parallelism, so there is no performance benefit.

  \begin{algorithm}
    \begin{algorithmic}[1]
  \Function{Scanner}{$(S \rightarrow \alpha \bullet T \beta, i, k)$}\label{ln:late-scan-start}
  \If{$W[k] = T$}
  \State $C \gets C \cup (S \rightarrow \alpha T \bullet \beta, i, k + 1)$
  \EndIf
  \EndFunction\label{ln:late-scan-end}
  \State
  \Function{Predictor}{$(S \rightarrow \alpha \bullet N \beta, i, k)$}\label{ln:late-pred-start}
  \tikzmark{late-pred-atomic-t}
  \State $Q[(N, k)] \gets Q[(N, k)] \cup (S \rightarrow \alpha \bullet N \beta, i, k)$\label{ln:late-pred-atomic-add}
  
  \If{$|Q[(N, k)]| = 1$}\label{ln:late-pred-atomic-if} // If $Q[(N,k]$ was previously empty \tikzmark{late-pred-atomic-b}\tikzmark{late-pred-atomic-r}
  \For{$N \rightarrow \gamma \in G$}
  \State $C \gets C \cup (N \rightarrow \bullet \gamma, k, k)$
  \EndFor
  \EndIf
  \For{$j \in P[(N, k)]$}
  \State $C \gets C \cup (S \rightarrow \alpha N \bullet \beta, i, j)$
  \EndFor
  \EndFunction\label{ln:late-pred-end}
  \State
  \Function{Completer}{$(S \rightarrow \alpha \bullet, i, k)$}\label{ln:late-cmp-start}\tikzmark{late-comp-atomic-r}
  \tikzmark{late-comp-atomic-t}
  \If{$(S, i, k) \not\in D$}\label{ln:late-comp-atomic-if}
  \State $D \gets D \cup (S, i, k)$\label{ln:late-comp-atomic-add}\tikzmark{late-comp-atomic-b}
  \State $P[(S, i)] \gets P[(S, i)] \cup k$
  \For{$(T \rightarrow \beta \bullet S \gamma, j, i) \in Q[(S, i)]$}
  \State $C \gets C \cup (T \rightarrow \beta S \bullet \gamma, j, k)$
  \EndFor
  \EndIf
  \EndFunction\label{ln:late-cmp-end}
  \State
  \Function{Earley}{$G, W$}
  \State Let $C$ be the Earley chart set. \label{ln:late-chart}
  \State Let $D$ be the completed set. \label{ln:late-completed}
  \State Let $Q$ be the requests map. \label{ln:late-requests}
  \State Let $P$ be the replies map. \label{ln:late-replies}
  \For{$\text{START} \rightarrow \beta \in G$}\label{ln:late-init-start}
  \State $C \gets C \cup (\text{START} \rightarrow \bullet \beta, 0, 0)$
  \EndFor\label{ln:late-init-end}
  \ForAll{States $s \in C$}
  \If{$s = (S \rightarrow \alpha \bullet, i)$} \Call{Completer}{$s$}
  \ElsIf{$s = (S \rightarrow \alpha \bullet T \beta, i)$ and $T$ is a
    non-terminal} \Call{Predictor}{$s$}
  \Else{\ \Call{Scanner}{$s$}}
  \EndIf
  \EndFor
  \State \Return $(\text{START} \rightarrow \alpha \bullet, 0, |W|) \in C$
  \EndFunction
\end{algorithmic}
\AddNote{late-pred-atomic-t}{late-pred-atomic-b}{late-pred-atomic-r}{Atomic for asynchronous execution}
\AddNote{late-comp-atomic-t}{late-comp-atomic-b}{late-comp-atomic-r}{Atomic for asynchronous execution}


    \caption{The serial LATE algorithm for parsing CFGs.}
    \label{alg:late}
  \end{algorithm}

\subsection{Serial LATE Parsing}
\label{sec:late}
  Earley parsing is difficult to parallelize because we cannot run the
completer on a finished Earley item originating at position $i$ and producing
some symbol $S$ until we have added all of the items at current position $i$
that will use $S$ as their next symbol. If we process our finished item too
soon, the items that use $S$ as their next symbol will not be advanced since
the predict step cannot advance rules.  This limitation forces the processing
order of the Earley items in increasing order of current location.

  This same limitation complicates Earley parsing of $\epsilon$ symbols (which
represent empty strings). Suppose we are about to use the Earley algorithm to
process the item $(S \to \bullet N N, i, k)$ and that there exists a
rule $N \to \epsilon$. Prediction adds $(N \to \bullet \epsilon, k, k)$
to the chart, then scanning adds $(N \to \epsilon \bullet, k, k)$.
Completion of $(N \to \epsilon \bullet, k, k)$ advances the original item
once, adding $(S \to N \bullet N, i, k)$ to the chart, but prediction
on $(S \to N \bullet N, i, k)$ will only add $(N \to \bullet
\epsilon, k, k)$, which has been added before.  Since the predict step cannot
advance rules, $(S \to N \bullet N, i, k)$ will never be advanced
again.

  There are special purpose fixes for the case of processing $\epsilon$
correctly, but we can actually solve both of these problems at the same time in
a more general way by adding three new data structures that allow the predict
step to advance Earley items. We call our modified algorithm the LATE
algorithm. The LATE parser produces the same chart as the Earley parser, but
tasks can be completed in parallel.  That LATE can also handle $\epsilon$ rules with only a minimal modification to the scan step came as a surprise to the authors, as we had limited our focus to asynchronous execution.

  The Earley parser stores a set for each input position $k$ filled with items of the form $(S \to \alpha \bullet \beta, i)$.  For simplicity, the LATE parser stores a global chart (line \ref{ln:late-chart}) with items of the form $(S \to \alpha \bullet \beta, i, k)$.
The LATE parser also stores three additional data structures:
\begin{itemize}
\item \emph{The requests map:} When the predict step processes an item $(S \to \alpha \bullet N \beta, i, k)$, where $N$ is a nonterminal, we say that $N$ has been \emph{requested} at input position $k$ by this item. The requests data structure (line \ref{ln:late-requests}) maps tuples $(N, k)$ of a symbol and input position to a list of items which requested the symbol $N$ at location $k$.
\item \emph{The replies map:} When the parse of a symbol is completed, we say that this is a \emph{reply}. The replies data structure (line \ref{ln:late-replies}) maps tuples $(N, i)$ of a symbol and an input position to a list of positions $k$ where $N$ has been parsed starting at $i$ and ending at $k$.
\item \emph{The completed set:} To avoid redundant append operations to reply lists, the completed set (line \ref{ln:late-completed}) stores tuples $(N, i, k)$ representing completed parses of nonterminals $N$ originating at position $i$ and ending at position $k$.
\end{itemize}

The scanner, predictor, and completer in the LATE algorithm are slightly modified to use these new data structures:
\begin{itemize}
\item \textsc{Scanner}: Like the Earley parser, the LATE parser only processes states once. The scan step is the same as in the Earley algorithm.

\item \textsc{Predictor}: The LATE predict step adds its state to the requests data structure, and adds advanced copies of this state to each position in the replies data structure. If no requests have yet been made for a nonterminal, then all rules which can parse the nonterminal are added to the current position.

\item \textsc{Completer}: The LATE complete step first checks if its completed parse of a symbol is in the completed set. If it is, the complete step can stop early. Otherwise, the symbol, origin, and current position tuple are added to the completed set. Next, we add the current (finished) location to the replies list at the origin of the completed symbol, then advance all states in the requests list at the origin of the completed symbol to the current position.
\end{itemize}

\subsubsection{Equivalence of Earley and LATE}
It is not initially obvious that the LATE algorithm will create the same chart as the Earley algorithm. We can prove that the LATE algorithm creates the same chart by induction. We will refer to a set of items which must be present in the chart for the new item to be added by a rule as \emph{parents} of the item. An item may have many parent sets. 

	In both algorithms, items of the form $(N \to \bullet \gamma, k, k)$, where $N$ is a nonterminal and $S \to \gamma$ is a rule, are formed from items of the form $(S \to \alpha \bullet N \beta, i, k)$ in predict steps and items of the form $(S \to \alpha T \bullet \beta, i, k + 1)$, where $T$ is a terminal starting at position $k$, are formed from items of the form $(S \to \alpha \bullet T \beta, i, k)$ in scan steps. The only other produced item forms are $(S \to \alpha N \bullet \beta, i, k)$, where $N$ is a nonterminal, which are produced when there exist $(N \to \gamma \bullet, j, k)$ and $(S \to \alpha \bullet N \beta, i, j)$ items in the chart. In the Earley algorithm, we can guarantee that $(S \to \alpha \bullet N \beta, i, j)$ is processed before $(N \to \gamma \bullet, j, k)$, so processing the later parent adds the child to the chart. In the LATE algorithm, there is no guarantee on the order of execution, but whichever parent is processed last will add the child to the chart.
    
    Since the charts are initialized to the same values, the sets of parents are the same, and both algorithms will always add a child if its parents have been added, the final charts will be the same.

  \subsection{Parallel LATE Parsing}
  \label{sec:late-parallel}
  We can run the LATE algorithm in parallel by processing items on a single shared work queue and updating the requests, replies, and completed sets atomically. We must modify lines (\ref{ln:late-pred-atomic-if} -- \ref{ln:late-pred-atomic-add}) to occur atomically, so that we only take the if-branch if the item was not already in the set when we atomically added the item. We must perform the same transformation on lines (\ref{ln:late-comp-atomic-if} -- \ref{ln:late-comp-atomic-add}). LATE has two subtle yet critical orders of execution. First, states must be added to the requests map before the replies map is traversed in the predict step. Second, positions must be added to the replies map before the requests map is traversed in a complete step. Race conditions in LATE might result in duplicate additions to the chart set (which are inconsequential), but never result in missed additions to the chart set.

\section{Testing}
We created three datasets for testing and evaluation in order to demonstrate the behavior of the algorithms on various types of inputs. In order to ensure that our implementations were correct (and not, say, accidentally saving work by incorrectly skipping parts of the algorithm or otherwise producing erroneous results), we implemented a reference version of the Earley algorithm in Python, and verified for each test set that the Earley charts were the same across all algorithms. We also checked the output of specific handcrafted examples that were known to be either valid or invalid parses.

\subsection{Arithmetic Grammar}
\label{sec:grammar-arith}
Our first dataset is an arithmetic grammar, in which valid sentences are arithmetic expressions such as \texttt{5 + 6 * 3}. This grammar is very small, but highly ambiguous, because expressions can be parsed in many ways depending on the order of operations. For example, \texttt{5 + 6 * 3} can be interpreted as either \texttt{(5 + 6) * 3} or \texttt{5 + (6 * 3)}. We wanted to explore the impact of grammar ambiguity on the parallelism of the LATE algorithm, so we derived increasingly ambiguous grammars from this seed grammar by replicating nonterminals. For example, instead of the nonterminal FACTOR, we would create the two identical nonterminals FACTOR0 and FACTOR1, either of which can be chosen arbitrarily in a valid parse. This approach allowed us to control the amount of ambiguity so that different levels of ambiguity could be compared. We also generated a synthetic corpus for this grammar.

\begin{figure}[!htb]
\begin{verbatim}
START     SUM
SUM       PROD
SUM       SUM + PROD
SUM       SUM - PROD
PROD      PROD * FACTOR
PROD      PROD / FACTOR
PROD      FACTOR
FACTOR    [ SUM ]
FACTOR    0
FACTOR    1
FACTOR    2
FACTOR    3
FACTOR    4
FACTOR    5
FACTOR    6
FACTOR    7
FACTOR    8
FACTOR    9
\end{verbatim}
  \caption{The seed arithmetic grammar.}
\end{figure}

\subsection{Java Grammar}
\label{sec:grammar-java}
Our second dataset is a grammar based on Java, taken from Oracle Corporation's grammar in the Java SE 7 language specification\footnote{\url{https://docs.oracle.com/javase/specs/jls/se7/html/jls-18.html}}. The Java language grammar is representative of one of the major use cases for CFG parsers, parsing programming languages. This grammar is fairly large and is not very ambiguous; there are very few valid parses even for large Java files.

Extracting a usable grammar from the Java specification required significant processing. For example, the syntax specifies that ``\{x\} denotes zero or more occurrences of x'', but does not distinguish between curly braces used for this purpose and literal curly brace characters used in Java code, so several of the syntax rules were ambiguous. In addition, descriptions of this nature needed to be converted to formal syntax rules: for example, "$XOrMore \to \text{at least one occurrence of x}$" would be converted to the rules "$XOrMore \to \text{x}$" and "$XOrMore \to \text{x}\ XOrMore$". We also removed $\epsilon$-rules ($x \to \epsilon$), since our Earley parser implementation would require significant modification to handle them.

The corpus for this grammar is an example Java file from ElasticSearch, one of the most popular Java repositories on GitHub\footnote{\url{https://github.com/elastic/elasticsearch/blob/master/client/benchmark/src/main/java/org/elasticsearch/client/benchmark/AbstractBenchmark.java}}. Since this was a newer Java version than the grammar we used, it required some manual processing to remove unsupported symbols.

\begin{figure}
\begin{verbatim}
START                            CompilationUnit
Identifier                       IDENTIFIER
Many.Identifier60                . Identifier Many.Identifier60
Many.Identifier60                . Identifier
QualifiedIdentifier              Identifier Many.Identifier60
QualifiedIdentifier              Identifier
Many,QualifiedIdentifier35       , QualifiedIdentifier Many,QualifiedIdentifier35
Many,QualifiedIdentifier35       , QualifiedIdentifier
QualifiedIdentifierList          QualifiedIdentifier Many,QualifiedIdentifier35
QualifiedIdentifierList          QualifiedIdentifier
MaybeAnnotations56               Annotations
ManyImportDeclaration88          ImportDeclaration ManyImportDeclaration88
ManyImportDeclaration88          ImportDeclaration
ManyTypeDeclaration19            TypeDeclaration ManyTypeDeclaration19
ManyTypeDeclaration19            TypeDeclaration
CompilationUnit                  MaybeMaybeAnnotations56packageQualifiedIdentifier;72 ManyImportDeclaration88 ManyTypeDeclaration19
Maybestatic71                    static
Many.QualifiedIdentifier1        . QualifiedIdentifier Many.QualifiedIdentifier1
Many.QualifiedIdentifier1        . QualifiedIdentifier
Maybe.*55                        . *
ImportDeclaration                import Maybestatic71 QualifiedIdentifier Many.QualifiedIdentifier1 Maybe.*55 ;
ImportDeclaration                import QualifiedIdentifier Many.QualifiedIdentifier1 Maybe.*55 ;
ImportDeclaration                import Maybestatic71 QualifiedIdentifier Maybe.*55 ;
ImportDeclaration                import QualifiedIdentifier Maybe.*55 ;
ImportDeclaration                import Maybestatic71 QualifiedIdentifier Many.QualifiedIdentifier1 ;
ImportDeclaration                import QualifiedIdentifier Many.QualifiedIdentifier1 ;
ImportDeclaration                import Maybestatic71 QualifiedIdentifier ;
ImportDeclaration                import QualifiedIdentifier ;
TypeDeclaration                  ClassOrInterfaceDeclaration
TypeDeclaration                  ;
ManyModifier65                   Modifier ManyModifier65
ManyModifier65                   Modifier
ClassOrInterfaceDeclaration      ManyModifier93 InterfaceDeclaration
ClassOrInterfaceDeclaration      InterfaceDeclaration
ClassDeclaration                 NormalClassDeclaration
ClassDeclaration                 EnumDeclaration
\end{verbatim}
  \caption{A portion of the Java grammar.}
\end{figure}

\subsection{English Grammar}
\label{sec:grammar-english}
For our third dataset, we decided to use our Grammy award-winning\footnote{The MIT 6.863 Grammy Awards, that is!} English sentence grammar. Our grammar is quite large, but the valid sentences are usually small. We therefore created a grammar which searches for valid sentences in a paragraph formed by concatenating many valid sentences, matching each valid English sentence in a corpus exactly once.
This grammar represents an NLP task: extracting sentences from a corpus. It
could be used for extracting sentences in the absence of punctuation, or for
detecting missing punctuation.
If $S_{English}$ is our English grammar and $S_{Wild}$ matches every terminal in our English grammar, then our third test grammar is:
\begin{align*}
	S &\to W S_{English}\\
    S &\to W S_{English} W\\
    S &\to S_{English} W\\
    W &\to W S_{Wild}\\
    W &\to S_{Wild}
\end{align*}
That is, a valid parse $S$ consists of a valid sentence (from $S_{English}$) surrounded by any number of terminals. Therefore, when this grammar is run on a corpus, each parse tree represents a particular sentence that was found, and each sentence is found in a particular parse tree; the number of parses is the number of distinct sentences.

The corpus for this dataset was a paragraph formed from the set of adversarial sentences from the CGW project, restricted to those sentences that were parsed by our S1 grammar.


\section{Evaluation}

In this section, we show that our modifications to the Earley algorithm allow it to scale across many cores, significantly increasing its performance when parsing ambiguous grammars.

\subsection{Implementation}
\label{sec:impl}

We implemented the serial Earley algorithm, and serial and parallel versions of
the LATE algorithm in C++\footnote{We also implemented a parallel version of the
  Earley algorithm, but it performed strictly worse than the serial version, so
  we do not include results for it.}\footnote{Our code and data is available at
  \url{https://github.com/jfeser/earley}.}. The serial Earley and LATE
implementations use the same code and data structures wherever possible, so we
can compare only the algorithmic differences. The parallel LATE implementation is based on Threading Building Blocks (TBB)~\cite{tbb}, a library for task parallelism. The parallel and serial versions use thread-safe and non-thread-safe variants, respectively, of the same TBB data structures.

\subsection{Experiments}
\label{sec:exp}

We tested the performance of our algorithm on our three datasets: a grammar of
arithmetic expressions (Sec.~\ref{sec:grammar-arith}), a Java
grammar(Sec.~\ref{sec:grammar-java}), and a grammar of English sentences
(Sec.~\ref{sec:grammar-english}).

For each grammar, we ran our serial Earley, serial LATE, and parallel LATE
implementations, running the parallel implementation with a varying number of
cores. We ran these experiments on an Intel Xeon E5-2470V2 with 10 physical
cores and 20 threads.

We tested four hypotheses about our algorithms:
\begin{enumerate}
  \item Both the serial and parallel LATE algorithms are faster than the serial
    Earley algorithm.
  \item The LATE algorithm exhibits \emph{strong scaling}, where an input
    can be processed in less time by adding more processors.
  \item The LATE algorithm exhibits \emph{weak scaling}, where an input with $n$
    units of work can be processed by $n$ processors in the time that one
    processor can perform one unit of work.
  \item The LATE algorithm scales better on ambiguous grammars.
\end{enumerate}

\subsection{Metrics}
	We used several metrics to judge our parallel implementations. The first, and simplest metric, is the \emph{runtime}. We measure the runtime $t_A$ of a parser $A$ as the wall clock time it takes from the moment when the parser has started to produce the chart to the moment when the chart is complete. This does not include the data structure setup time, or the time required to preprocess the grammar. Our runtimes are the result of 100 trials or enough trials to achieve a total time greater than 1 second, whichever happens first.
    
    The \emph{speedup} is a metric used to show how much faster one algorithm is with respect to another. The speedup $s_{A, B}$ of a parallel algorithm $A$ on some input with respect to $B$ is
    \[
    	s_{A, B} = \frac{t_B}{t_A}
    \]
    Where $t_A$ and $t_B$ are runtimes of each algorithm on the input.
    
    The \emph{efficiency} is a metric used to show how much work an algorithm is performing per element. Higher efficiency is better. The efficiency $e_A$ of a parallel parser $A$ using $p$ processors on an input which produces a chart with $n$ elements is
    \[
    	E_{A} = \frac{n}{pt_A}
    \]
    Note that if there are 10 cores and 20 threads, $p = 10$.

\subsection{Results}
\label{sec:res}

In this section we discuss the results of our experiments and comment on our
four hypotheses.

\begin{figure}
  \centering
  \begin{subfigure}[t]{0.48\textwidth}
    \centering
    \includegraphics[width=\textwidth]{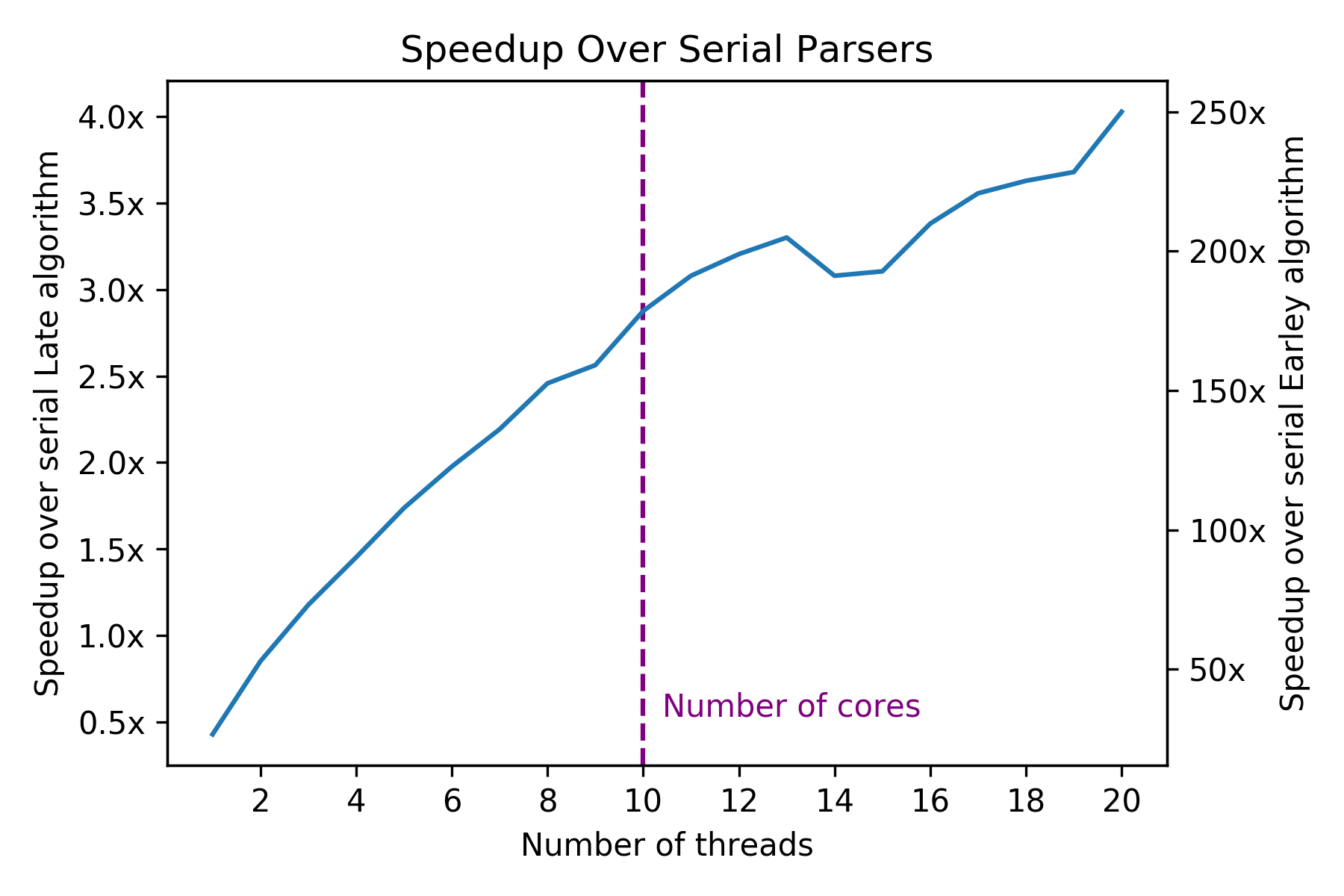}
    \caption{Parallel speedup on arithmetic grammar with 10 copies of each
      nonterminal.}\label{fig:speedup-arith}
  \end{subfigure}%
  \hspace{1em}\begin{subfigure}[t]{0.48\textwidth}
    \centering
    \includegraphics[width=\textwidth]{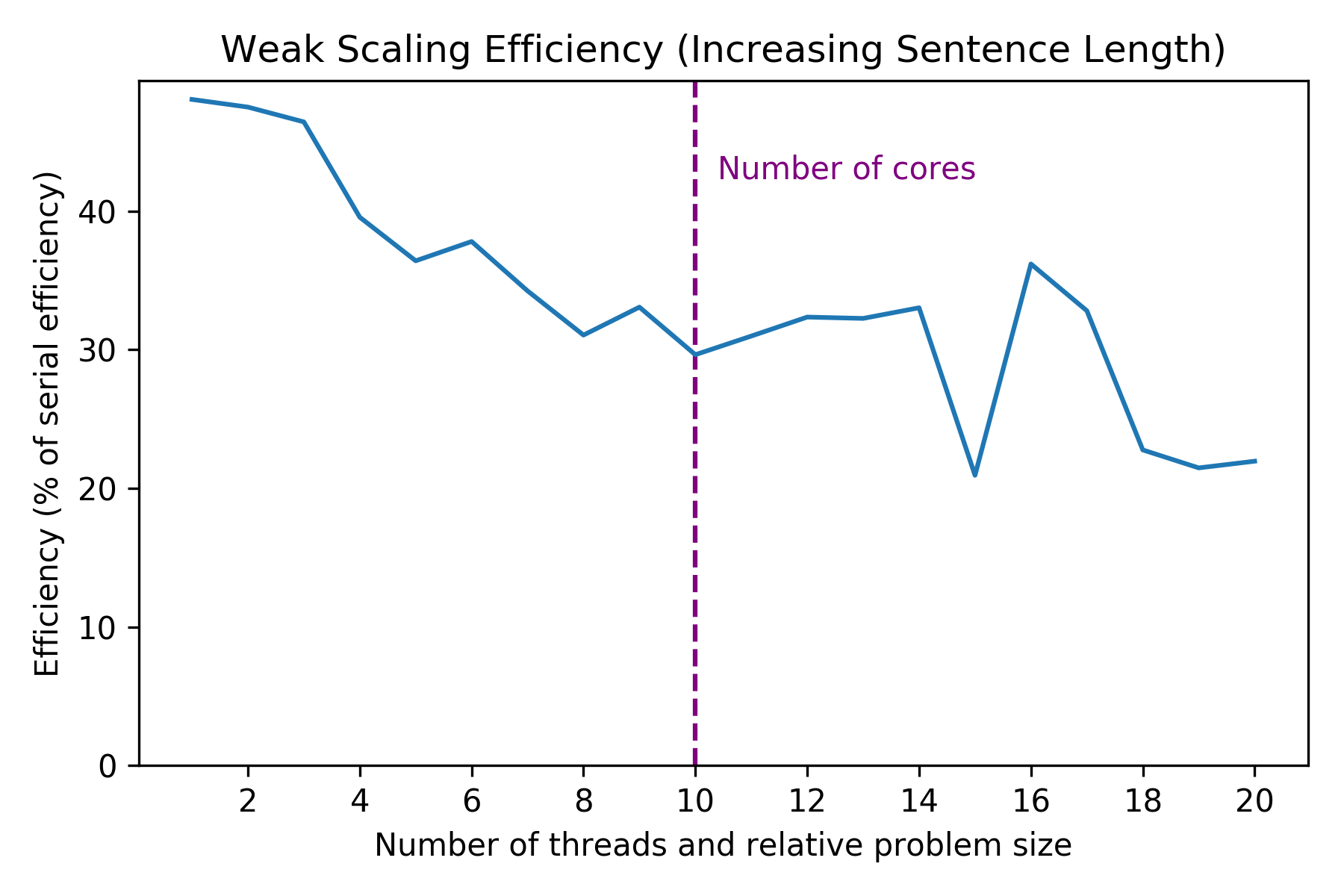}
    \caption{Weak scaling efficiency for arithmetic grammar with 10 copies of
      each nonterminal, expressed as a percentage of the serial efficiency.}\label{fig:weak-arith}
  \end{subfigure}
  \begin{subfigure}[t]{0.48\textwidth}
    \centering
    \includegraphics[width=\textwidth]{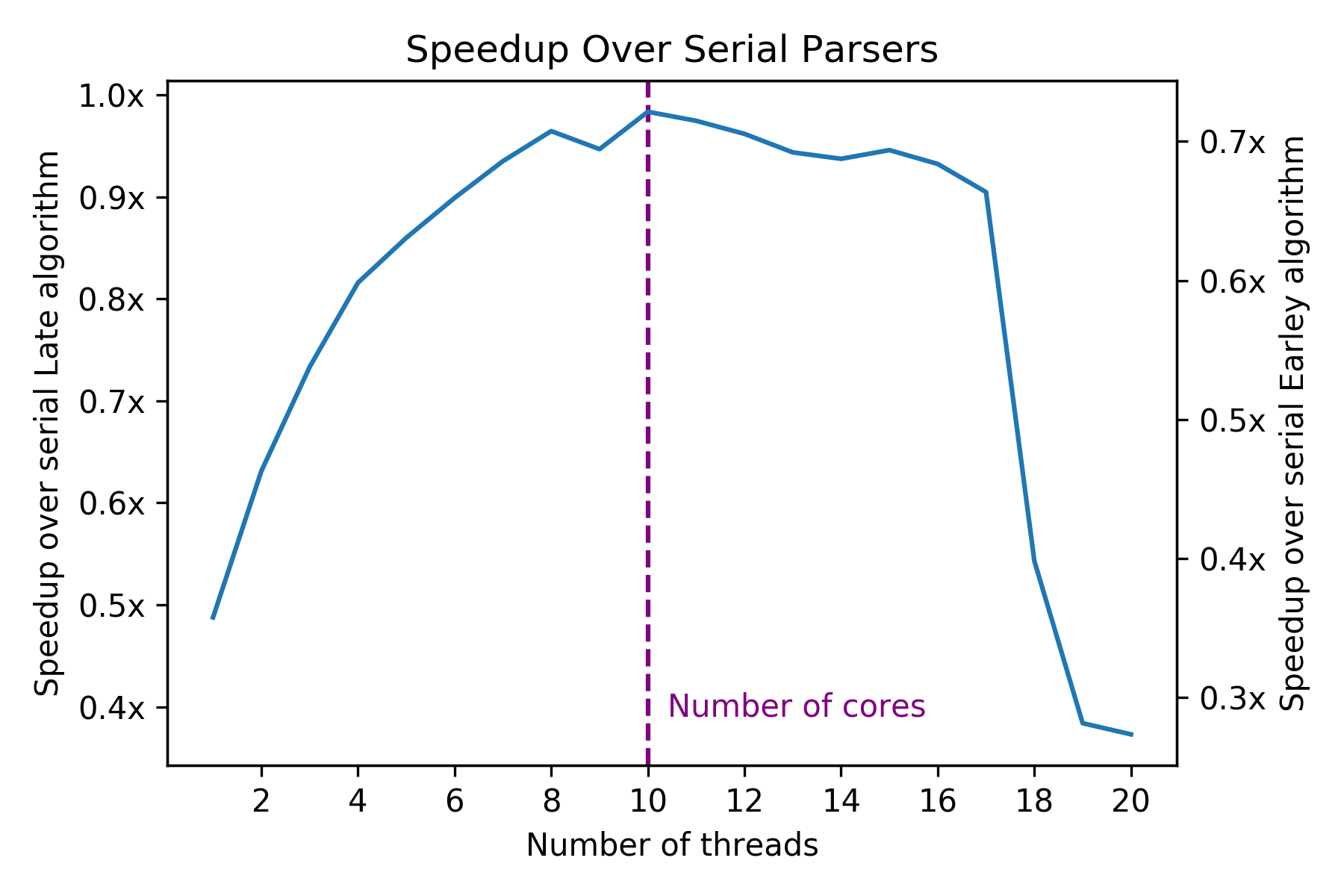}
    \caption{Parallel speedup on Java grammar.}\label{fig:speedup-java}
  \end{subfigure}%
  \hspace{1em}\begin{subfigure}[t]{0.48\textwidth}
    \centering
    \includegraphics[width=\textwidth]{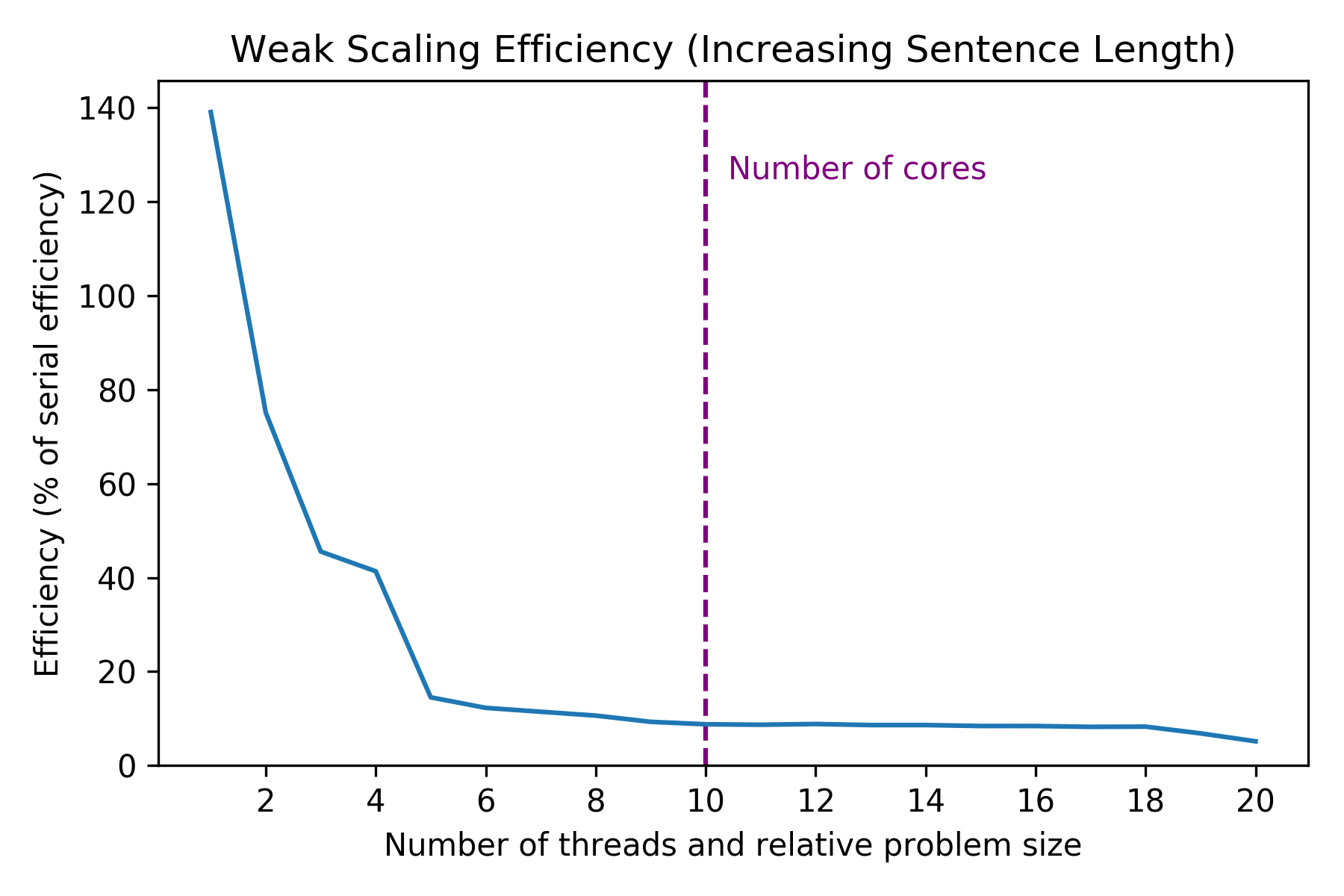}
    \caption{Weak scaling efficiency on Java grammar, expressed as a percentage of the serial efficiency.}\label{fig:weak-java}
  \end{subfigure}
  \begin{subfigure}[t]{0.48\textwidth}
    \centering
    \includegraphics[width=\textwidth]{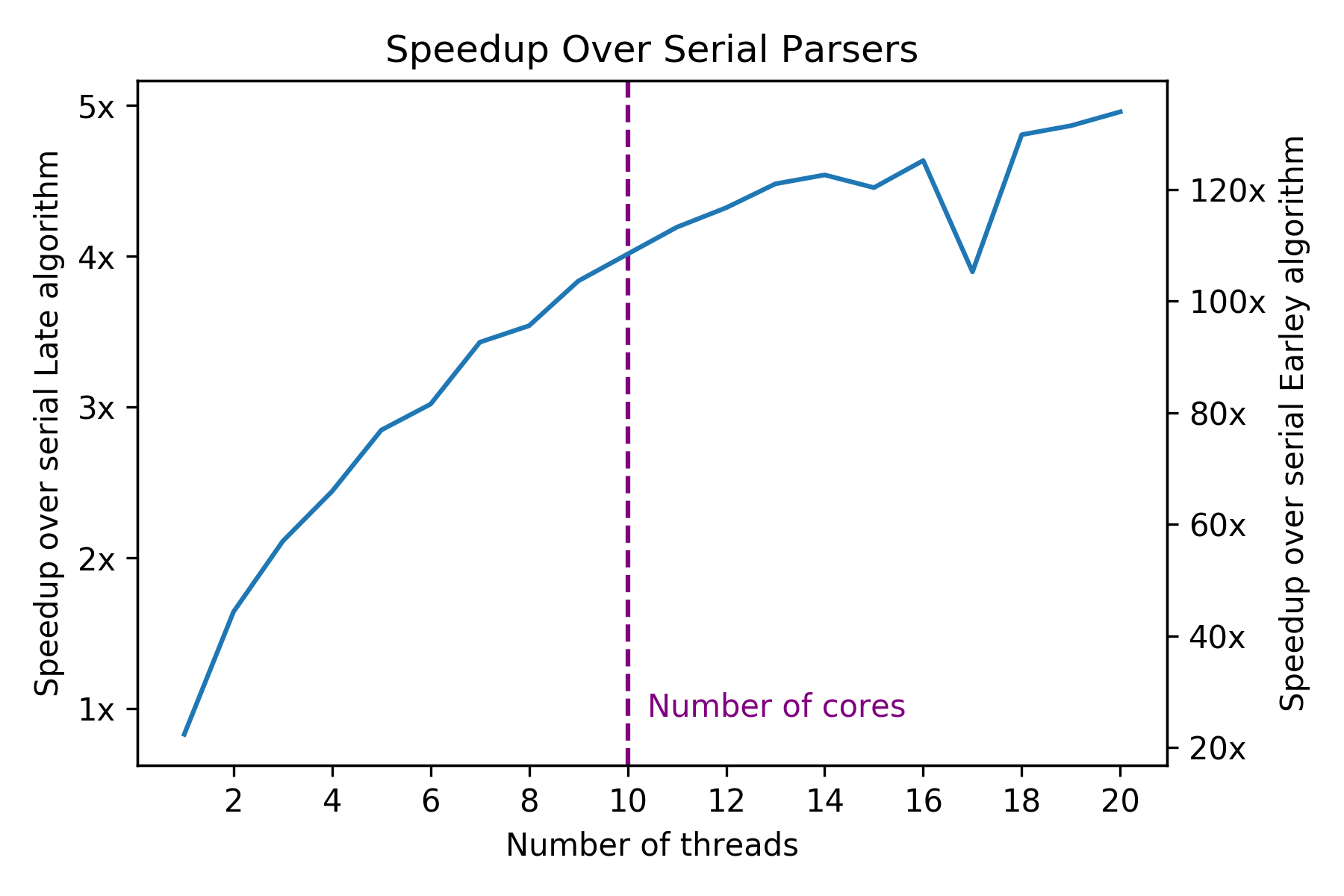}
    \caption{Parallel speedup on English sentence grammar.}\label{fig:speedup-cgw}
  \end{subfigure}%
  \hspace{1em}\begin{subfigure}[t]{0.48\textwidth}
    \centering
    \includegraphics[width=\textwidth]{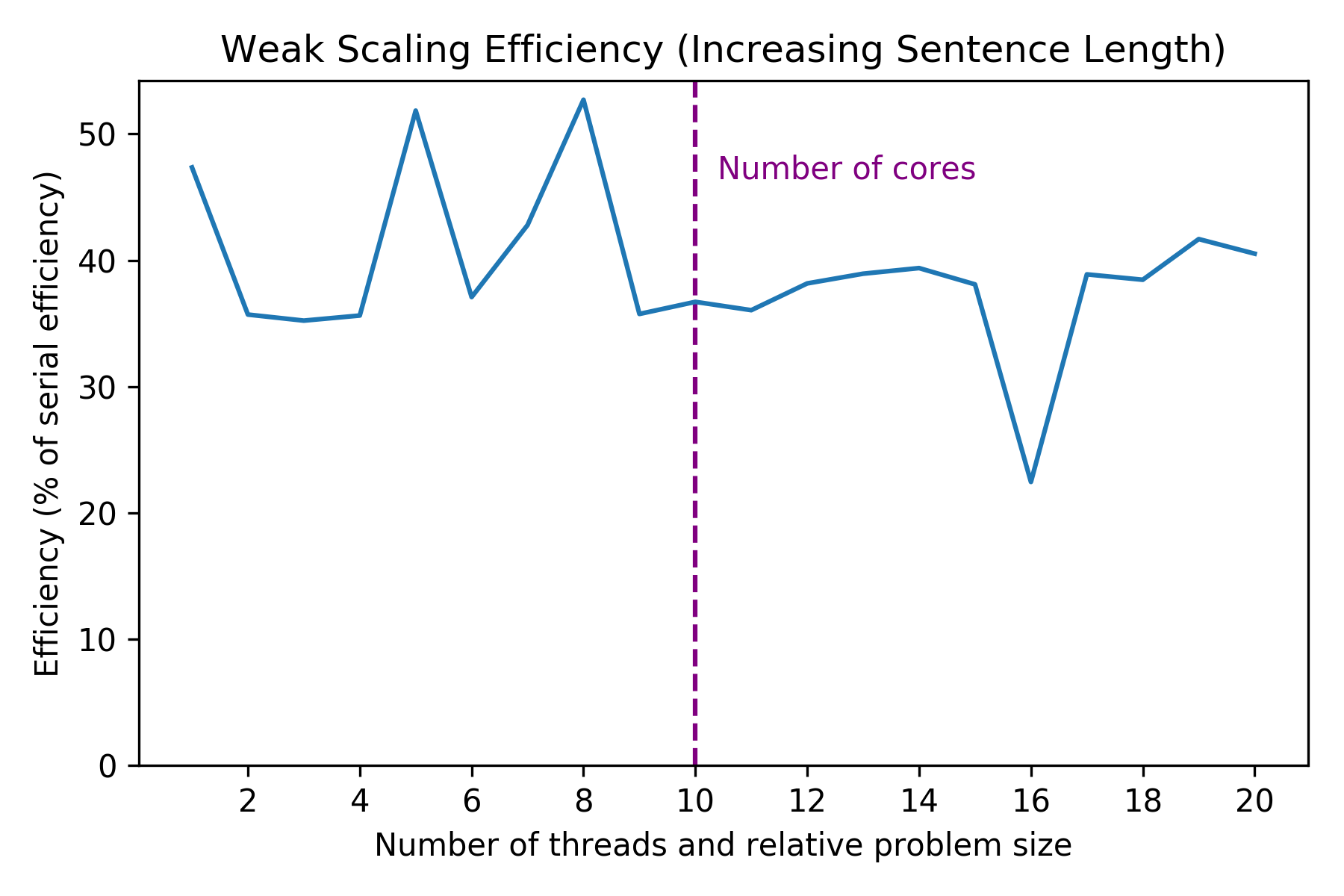}
    \caption{Weak scaling efficiency on English sentence grammar, expressed as a percentage of the serial efficiency.}\label{fig:weak-cgw}
  \end{subfigure}
  \caption{Parallel speedup for benchmark grammars.\label{fig:speedup}}
\end{figure}

\paragraph{Hypothesis 1: LATE is faster than Earley}

In Fig.~\ref{fig:speedup-arith} and Fig.~\ref{fig:speedup-cgw}, we can see that
the serial LATE algorithm is significantly more efficient than the serial Earley
algorithm for ambiguous grammars. We obtain 100x and 20x speedups respectively
from the algorithmic change alone. This is because the additional index
structures that we add allow the LATE algorithm to do much less work,
particularly in the \textsc{Predictor} and \textsc{Completer} functions.

The Java grammar (Fig.~\ref{fig:speedup-java}) is a worst case scenario for the
LATE algorithm. It has very little ambiguity, which means that the overhead of
the additional data structures in the serial LATE algorithm does not pay off.

We believe that Hypothesis 1 is true in some cases. In particular, LATE performs
better than Earley when the grammar is ambiguous.

\paragraph{Hypothesis 2: Strong Scaling}

Fig.~\ref{fig:speedup} shows the performance of the parallel LATE algorithm on
each grammar, scaled against the serial Earley and LATE algorithms. Parallelism
gains an additional 4-5x on 10 physical cores for the arithmetic and English
grammars, for an efficiency of 40-50\%.

When parsing the Java grammar (Fig.~\ref{fig:speedup-java}) the parallel LATE
implementation has to pay the cost of communication between threads as well as
the overhead of using thread-safe data structures. The lack of parallel work 

Hyperthreading~\cite{marr_hyper-threading_2002} has a mixed impact. It leads to
an additional 2x speedup on the arithmetic grammar benchmark but has little
positive effect on the other benchmarks. This is somewhat surprising, because
hyperthreading is designed to improve performance on workloads that frequently
wait on memory. Our implementations rely on complex data structures like hash
maps and hash sets which are unlikely to lead to efficient memory access
patterns. Hyperthreading should allow us to scavenge many of the cycles that we
waste waiting on loads. We believe that in the case of the arithmetic benchmark,
the parallel tasks are sufficiently independent, so hyperthreading is helpful.
In the other benchmarks we believe that communication costs reduce the benefits
of using additional threads.

The LATE algorithm exhibits reasonably strong scaling, so Hypothesis 2 is true,
but only for ambiguous grammars.

\begin{figure}
  \centering
  \begin{subfigure}[t]{0.48\textwidth}
    \centering
    \includegraphics[width=\textwidth]{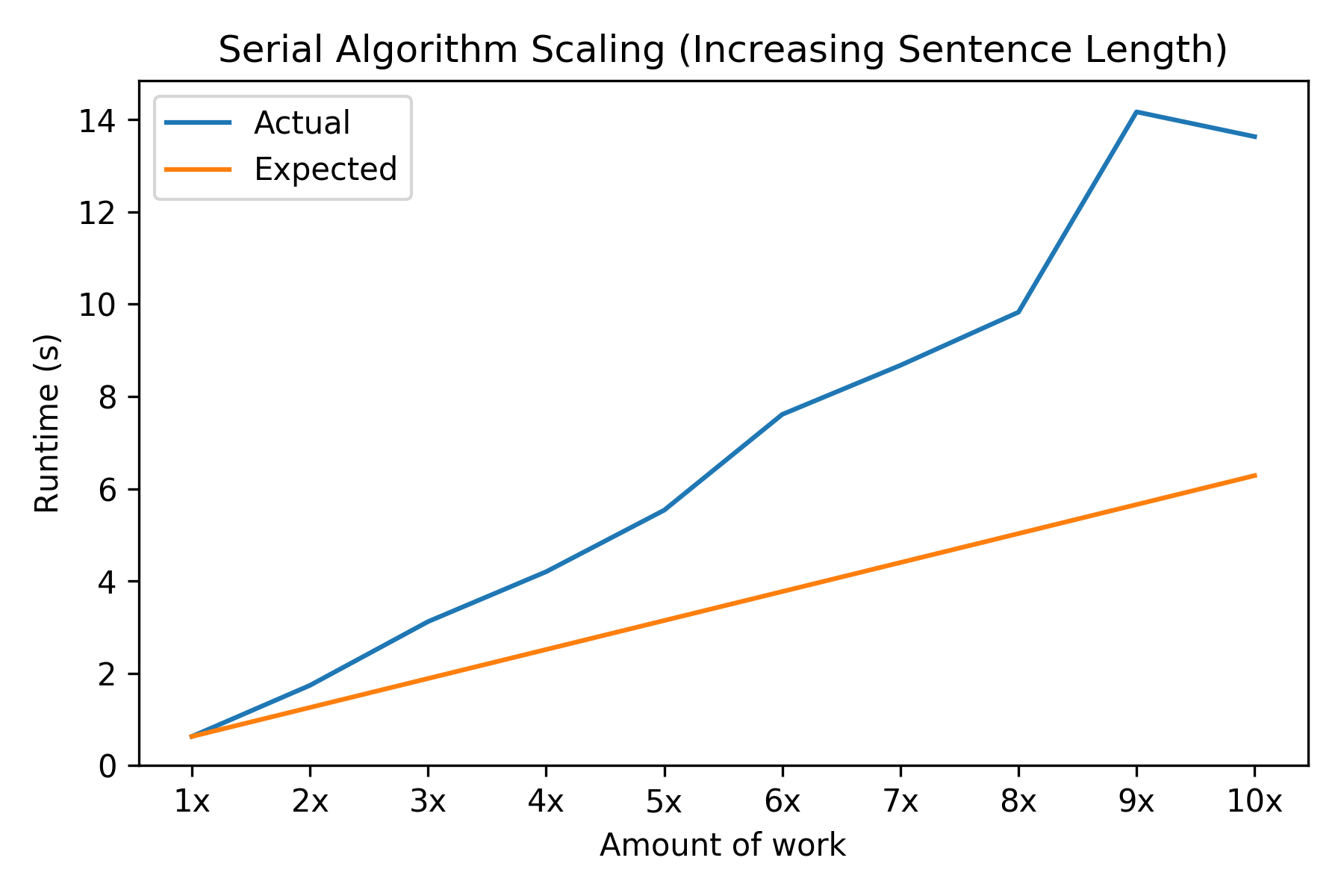}
    \caption{Serial scaling on arithmetic grammar with 10 copies of each
      nonterminal.}\label{fig:serial-scaling-arith}
  \end{subfigure}
  \begin{subfigure}[t]{0.48\textwidth}
    \centering
    \includegraphics[width=\textwidth]{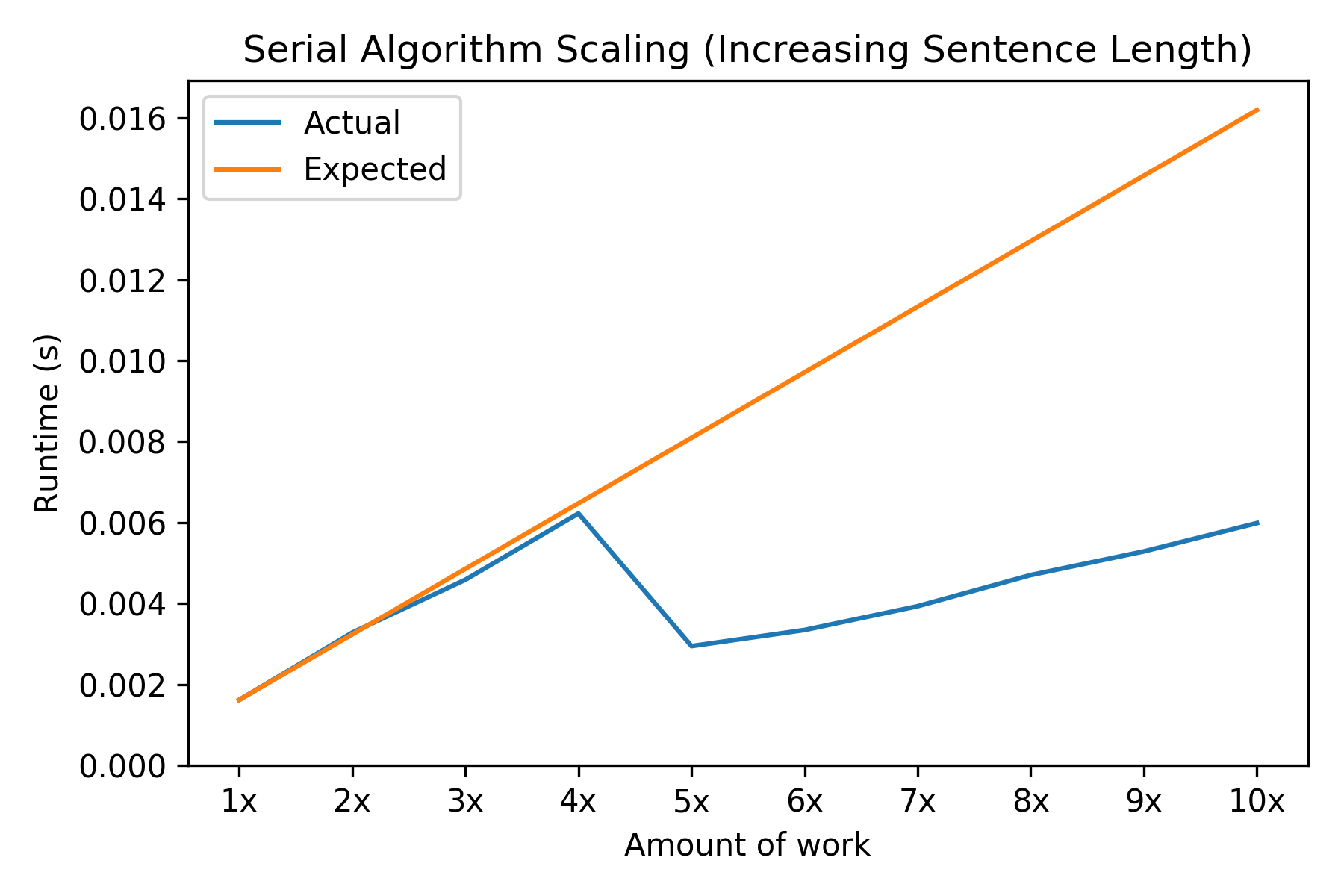}
    \caption{Serial scaling on Java grammar.}\label{fig:serial-scaling-java}
  \end{subfigure}
  \begin{subfigure}[t]{0.48\textwidth}
    \centering
    \includegraphics[width=\textwidth]{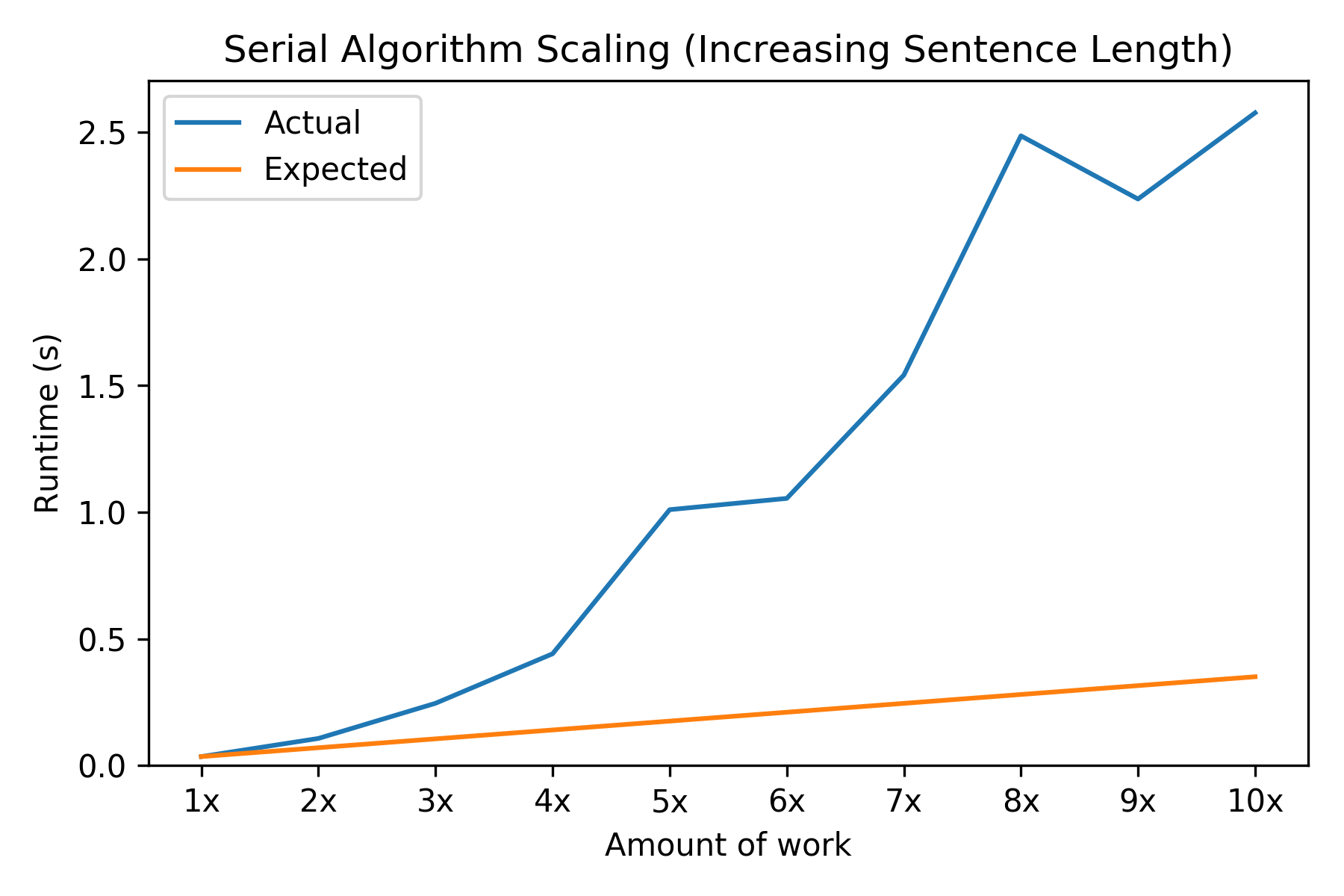}
    \caption{Serial scaling on English sentence grammar.}\label{fig:serial-scaling-cgw}
  \end{subfigure}
  \caption{Serial scaling on benchmark grammars. The expected time is computed
    by scaling the actual time for a problem of size one linearly.}
  \label{fig:serial-scaling}
\end{figure}

\paragraph{Hypothesis 3: Weak Scaling}

We ran into two problems when testing for weak scaling. 

The first problem we encountered is generating inputs which have the correct
amount of work. In general, weak scaling is tested by creating inputs with $n$
units of work for $n$ from 1 to the number of processors. Under weak scaling,
the time to run on input $k$ with $k$ processors should be the same as to run on
input 1 with 1 processor. In our case, inputs consist of both a grammar and a
sentence; the amount of work is given by the number of items in the Earley
chart, which is not directly related to either the size of the sentence or the
size of the grammar. To construct inputs that are properly scaled, we fixed a
grammar and an initial sentence, then used a binary search to select a prefix of
the sentence which has the correct number of items in its Earley chart.

The second problem is that the serial LATE algorithm does not seem to scale
linearly in the size of the Earley chart in all cases.
Fig.~\ref{fig:serial-scaling} shows the serial LATE algorithm along with an
extrapolated runtime, computed by scaling the runtime on the input of size one.
We believe that this nonlinear scaling is due to memory effects. It may occur
when we scale past the size of a cache. Unfortunately we did not have time to fully
investigate the cause. To correct for this effect, our weak scaling plots show
the percentage of the true serial efficiency for each problem size, rather than
scaling up the serial efficiency for the problem of size one.

Fig.~\ref{fig:weak-arith}, Fig.~\ref{fig:weak-java} and Fig.~\ref{fig:weak-cgw}
show the weak scaling behavior of the LATE algorithm. LATE scales reasonably
well on the arithmetic and the English grammars and poorly on the Java grammar,
so Hypothesis 3 is true for ambiguous grammars.

\begin{figure}
  \centering
  \includegraphics[width=0.6\textwidth]{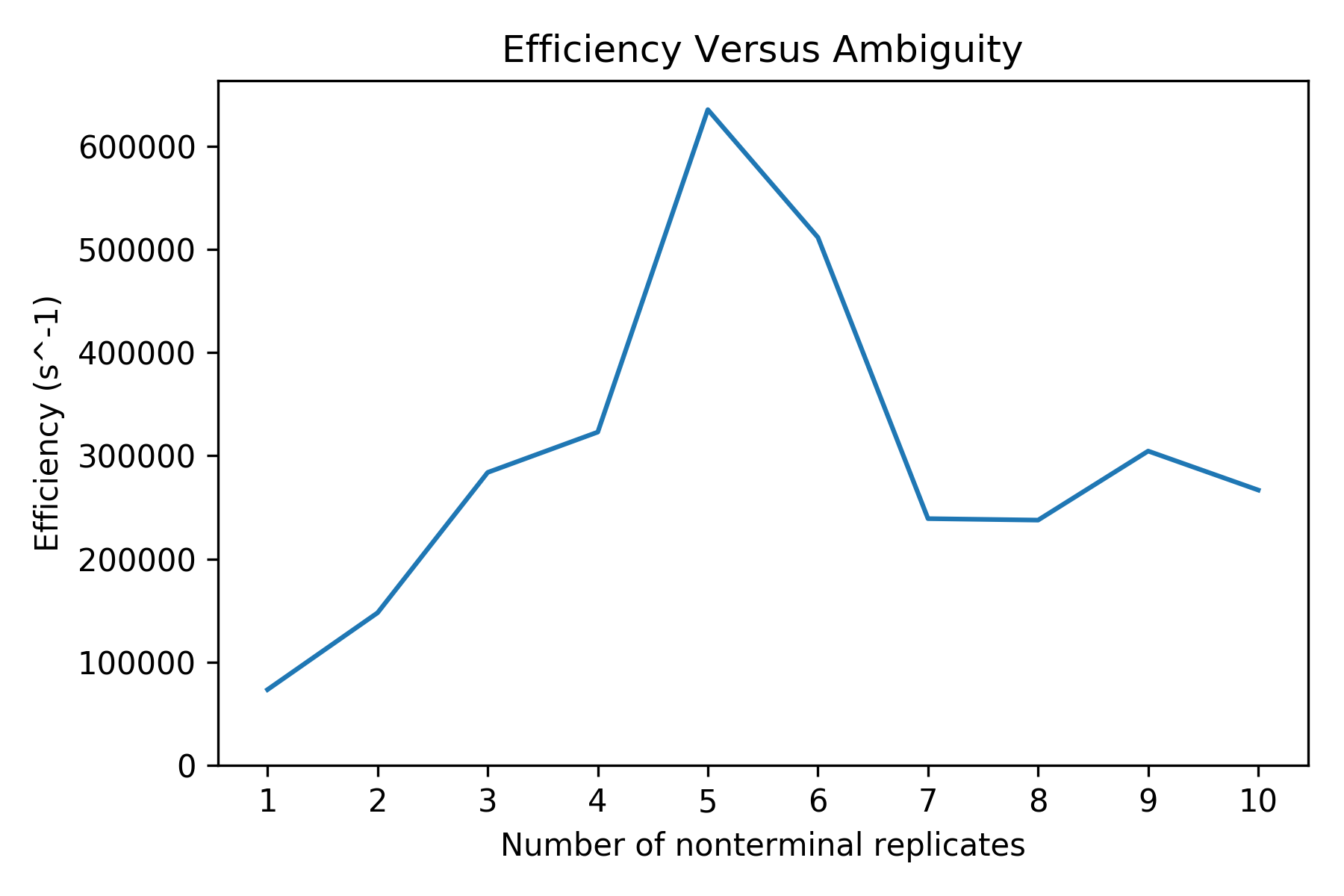}
  \caption{Items per second for increasingly ambiguous grammars. (Run with 20
    threads/10 cores).\label{fig:efficiency}}
\end{figure}

\paragraph{Hypothesis 4: LATE scales on ambiguous grammars}

Our strong and weak scaling results led us to look into the performance of the
LATE algorithm on grammars with varying ambiguities. To do so we generated
variants of the arithmetic grammar as described in Sec.~\ref{sec:grammar-arith}.
We used each, increasingly ambiguous grammar to parse a sentence and measured
the efficiency of the parser in each case. The number of cores and the size of
the sentence are kept constant. Fig.~\ref{fig:efficiency} shows the number of
Earley items processed per second as the ambiguity of the grammar increases. The
efficiency increases up to 4 nonterminal replicates, then levels off somewhat.
This chart shows that more ambiguous grammars have more exploitable parallelism,
but that the amount of exploitable parallelism in the grammar is not the
bottleneck for the algorithm. Therefore, Hypothesis 4 is true. The LATE
algorithm does scale better on ambiguous grammars, but there are other factors
that limit the available parallelism.


\section{Conclusions}

We have presented a modification of the Earley algorithm for parsing arbitrary context-free grammars that allows it to scale across multiple cores. Our algorithm performs particularly well when the grammar is very ambiguous, permitting many different parses of the input. This property makes our algorithm well suited for linguistics applications. However, if the grammar is unambiguous, or only has limited ambiguity, our algorithm scales poorly, and often does not overcome the parallel overhead. This limits its usefulness in parsing artificial languages, such as programming languages.

Our parallel implementation of the LATE algorithm assigns tasks to processors arbitrarily, and executes them in an arbitrary order. Because of the implementation of TBB, items added to a thread's work queue are most likely executed by that same thread, unless another thread has run out of work to do. This explains some of the difficulty in parallelizing efficient deterministic grammars. Future work includes the use of heuristics to help choose tasks in order to discover independent work for parallel processing. For example, in the beginning of the algorithm, selecting complete and scan tasks before predict tasks would help expose some of the work on later parts of the sentence. This would allow some threads to work on later parts of the input while other threads work on earlier parts. As another example, when threads need to steal work they might want to choose predict tasks, since these tasks usually spawn many more tasks than scan or complete tasks do.

The LATE algorithm may also extend to speculation-based parallelism in a similar manner to that proposed by Fowler et. al. \cite{fowler_parallel_2009}. Fowler's implementation required dividing the input into blocks based on the origin of items in order to correctly complete and scan. Since the LATE algorithm can execute the tasks in any order, it may be easier to implement speculative blocking using the LATE algorithm, where blocks are executed in parallel and the work within blocks are also executed in parallel using the LATE algorithm.

Our implementation may also admit serial speedups. \cite{mclean_faster_1996} introduces an optimization that improves the speed of Earley parsing. Their method combines LR(k) parsing, which is efficient due to its use of precomputed parse tables but can only recognize unambiguous context-free grammars, with Earley parsing, which is less efficient but can recognize ambiguous grammars. They achieve a 10-15x speedup over a conventional Earley parser when parsing C code. It may be possible to integrate this serial optimization with the LATE parser.

Although the datasets that we used for testing represent an array of different types of grammars, there are many interesting grammars and datasets that we did not test. A more thorough analysis could include real-world natural language datasets such as the Penn Treebank paired with an appropriate grammar (such as one trained using the Berkeley Parser), other computer languages such as HTML, or an array of many different corpora for the same grammar (e.g. several Java files written in different styles).

As a final note of future work, we could augment our implementation with a mechanism to reconstruct parse trees efficiently, and explore parallel parse-tree reconstruction.

Earley parsing has been ignored for practical parsing implementations as an inefficient algorithm, but we have shown that for ambiguous grammars, Earley parsing can be effectively parallelized. Highly ambiguous grammars are common in natural language processing, so Earley parsing (especially the LATE algorithm) may be a good parallel parser for these tasks.


\printbibliography
\end{document}